\documentclass[10pt,twocolumn,letterpaper]{article}

\usepackage[pagenumbers]{iccv} 

\definecolor{iccvblue}{rgb}{0.21,0.49,0.74}
\usepackage[pagebackref,breaklinks,colorlinks,allcolors=iccvblue]{hyperref}
\usepackage{multirow}
\usepackage{graphicx}
\def\paperID{8766} 
\def\confName{ICCV}
\def\confYear{2025}

\title{Seeing Through Deepfakes: A Human-Inspired Framework for Multi-Face Detection \thanks{This paper has been accepted by ICCV 2025.}}

\author{
Juan Hu, Shaojing Fan, Terence Sim\\
National University of Singapore\\
{\tt\small \{huj, fanshaojing, terence.sim\}@nus.edu.sg}
}

\begin{document}
\maketitle
\begin{abstract}
Multi-face deepfake videos are becoming increasingly prevalent, often appearing in natural social settings that challenge existing detection methods. Most current approaches excel at single-face detection but struggle in multi-face scenarios, due to a lack of awareness of crucial contextual cues. In this work, we develop a novel approach that leverages human cognition to analyze and defend against multi-face deepfake videos. Through a series of human studies, we systematically examine how people detect deepfake faces in social settings. Our quantitative analysis reveals four key cues humans rely on: scene-motion coherence, inter-face appearance compatibility, interpersonal gaze alignment, and face-body consistency. Guided by these insights, we introduce \textsf{HICOM}, a novel framework designed to detect every fake face in multi-face scenarios. Extensive experiments on benchmark datasets show that \textsf{HICOM} improves average accuracy by 3.3\% in in-dataset detection and 2.8\% under real-world perturbations. Moreover, it outperforms existing methods by 5.8\% on unseen datasets, demonstrating the generalization of human-inspired cues.  
\textsf{HICOM} further enhances interpretability by incorporating an LLM to provide human-readable explanations, making detection results more transparent and convincing. Our work sheds light on involving human factors to enhance defense against deepfakes.

\end{abstract}    
\section{Introduction}
\label{sec:intro}

The rapid rise of AI-generated content has made it easier to create and spread fake videos featuring multiple altered faces, increasing the risk of public manipulation and harm \cite{waseem2023deepfake}. Since humans naturally interact in groups, detecting fake faces in multi-face social settings is especially critical. For instance, a recent news report \cite{hongkong} revealed how fraudsters used fake faces of a CFO and employees in a video group meeting to deceive and defraud HK\$25 million. This case underscores the urgent need for detection methods that account for group dynamics and contextual cues to prevent such deceptive practices.

Accurately detecting every fake face within a video frame is crucial, as understanding a scene depends on the interplay among all faces present. For example, a deepfake video circulating on social media falsely depicts U.S. President Donald Trump, Vice President J.D. Vance, and Ukrainian President Volodymyr Zelenskyy engaging in a physical altercation inside the White House following their February 2025 dispute \cite{Trumpbbc}. The video, which gained traction on TikTok \cite{trumpZ}, misrepresents the nature of the confrontation. If only Trump's manipulated face is flagged as fake while Zelenskyy’s and Vance’s faces are mistakenly classified as genuine, viewers may still believe the fight occurred, albeit with Zelenskyy and Vance only. This highlights the need for comprehensive, frame-level complete multi-face detection\footnote{Frame-level complete multi-face detection considers a frame accurate only if every single face within the frame is correctly classified.} in social settings.



\begin{figure}[!t]
 \centering
\includegraphics[width=0.8\linewidth]{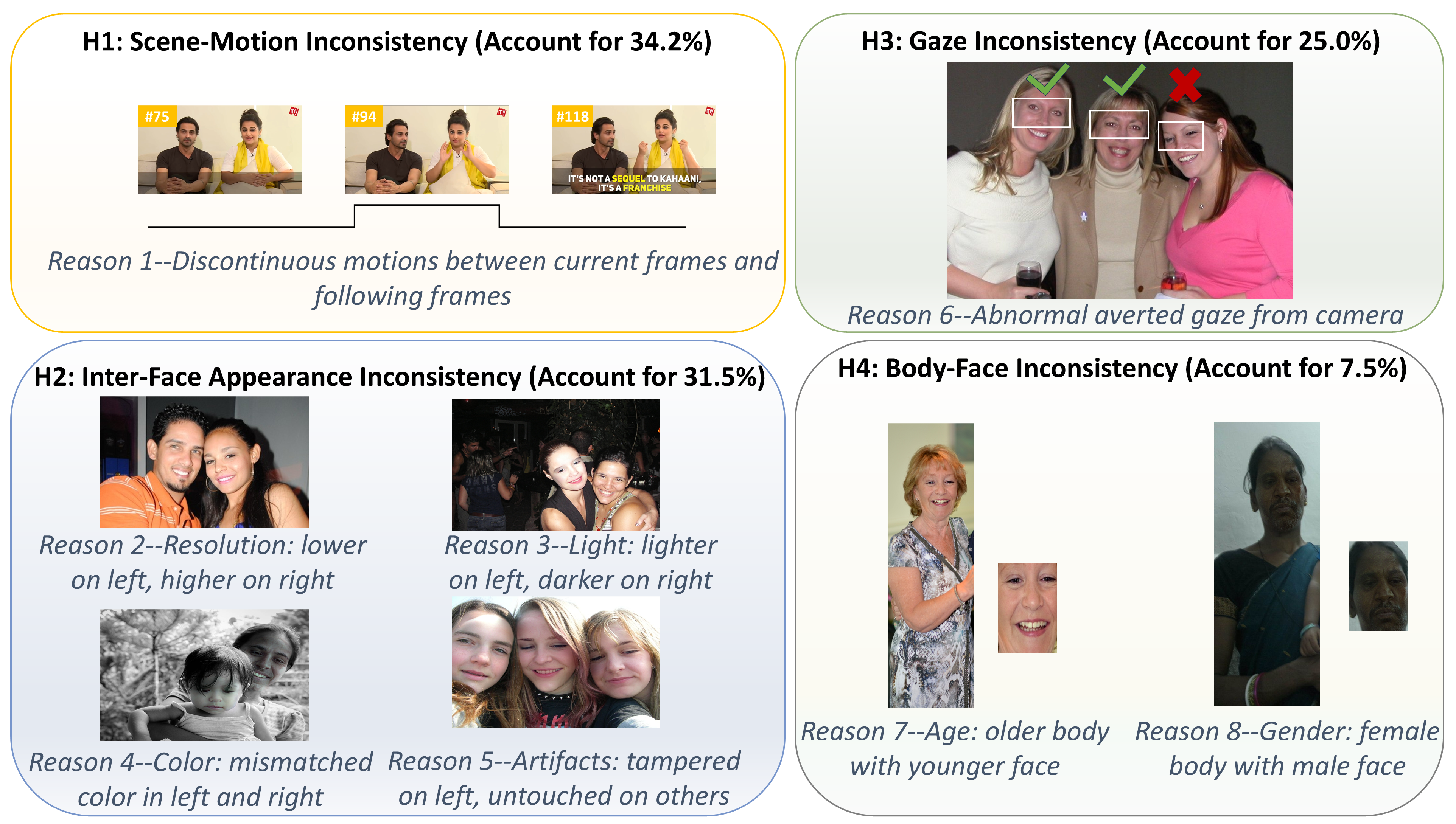}
\captionsetup{singlelinecheck=off} 
\caption{This work takes a novel human-centric approach to multi-face deepfake detection. We conducted a series of human studies to test four research hypotheses (H1-H4), identifying eight cues that humans rely on to detect deepfake faces. These cues later informed the design of our computational model, \textsf{HICOM}, which detects every deepfake face in multi-face scenarios.}
 \label{fig1-challenge}
 \vspace{-0.2CM}
\end{figure}

However, \textit{multi-face deepfake detection is an emerging scenario that is inherently more complex than traditional single-face detection, introducing new challenges.} Unlike single-face scenarios—where methods focus solely on individual facial characteristics—multi-face scenarios require understanding how multiple faces interact within the same scene.  Simply applying single-face detection methods independently to each face in a multi-face setting is insufficient because it ignores critical contextual information about the scene-motion relationships and interactions among faces \cite{lin2023exploiting,zhang2024comics}. In multi-face settings, faces often engage in visual interactions that introduce perceptual cues—such as temporal correlation, coherence in visual attributes, and scene consistency—that are essential for accurate interpretation. 

\noindent \textbf{Challenges.} \textit{The deepfake detection community currently lacks a well-established benchmark for frame-level complete multi-face detection.} Existing methods focus mainly on single-face detection \cite{wang2023noise,lin2024preserving,zhang2024comics,pellicer2024pudd,xu2023tall,ba2024exposing,tan2023deepfake,choi2024exploiting} and overlook essential contextual information, such as the scene relationships, interactions, and motion consistency among faces. This can result in misclassifying manipulated faces, which in turn may distort the overall interpretation of an event. Although a few multi-face detection methods exist \cite{zhou2021face,li2020sharp,ma2022accurate,lin2023exploiting,zhang2024comics,miao2024mixture}, they generally do not assess frame-level complete multi-face detection performance.  Moreover, these methods extract features based on the heuristics of individual researchers, limited to individual perceptions or assumptions. \textit{In multi-face deepfake scenarios, understanding how human cognition identifies deepfake cues in group settings remains an open research question.}

\noindent \textbf{Our approach and rationale.} In our work, we tackle the challenges of multi-face deepfake detection through a novel human-centric approach. Unlike previous methods that depend on off-the-shelf classifiers or heuristics shaped by individual researchers' perceptions, we focus on examining detection cues derived from crowdsourced human studies. This approach provides fresh insights into multi-face deepfake detection. By aggregating crowdsourced annotations, we identify multiple social contextual cues humans rely on for detecting deepfakes. Building on these cues, we propose \textsf{HICOM}, a framework designed to effectively detect multiple deepfake faces in social settings.


\textsf{HICOM} is grounded in a key rationale: rather than relying on off-the-shelf classifiers or individual heuristics, we base our model on human cognition. This is crucial because AI generation techniques evolve rapidly, and methods tailored to specific deepfake types often struggle to adapt to new variations. In contrast, human cognitive patterns remain stable, providing a reliable foundation for detection. Our rationale is supported by several studies in social science and neuroscience. First, research in social science shows that social context---capturing relationships within a group---plays a vital role in identifying inconsistencies \cite{gallagher2009understanding}. Additionally, neuropsychological studies reveal that the human visual system is highly attuned to face perception \cite{farah1998special,kanwisher2000domain,joslin2024double}. We hypothesize that this sensitivity gives humans a natural advantage in detecting deepfake faces, as they can instinctively recognize fake faces that don't fit within a social group \cite{gaither2016social}. Motivated by these insights, we explore how humans detect multi-face deepfakes in social settings and believe that incorporating human cognition can lead to effective and robust multi-face detection models.


As shown in Fig. \ref{fig1-challenge}, our human study reveals four key insights into the specific cues that frequently appear in multi-face deepfake scenarios, namely scene-motion coherence, inter-face appearance compatibility, interpersonal gaze alignment, and face-body consistency. 
Leveraging human insights, we identify key cognitive strategies used to detect fake faces in group contexts.  These insights inform the development of our \underline{h}uman-\underline{i}nspired, \underline{co}ntext-aware, \underline{m}ulti-face deepfake detection framework, named \textsf{HICOM}. The framework consists of four modules:  scene-motion module ({M1}), inter-face appearance module ({M2}), gaze module ({M3}), and body-face module ({M4}), with the weights of each module motivated by our human studies.  

\noindent Our contributions are as follows.

\begin{itemize}

\item \textbf{Human-inspired deepfake detection.} We pioneer the use of human studies to explore contextual features for multi-face deepfake detection in social settings. Our work introduces a novel analytical perspective and identifies key detection cues through a series of empirical studies. Insights from these studies inform the design of \textsf{HICOM}, which is tailored to align with how humans naturally detect deepfakes. This approach enhances detection interpretability and strengthens the persuasiveness of the results.


\item \textbf{Human cognition on deepfake perception. } We examine human cognitive patterns in identifying fake faces in multi-face deepfake videos, identifying four key factors humans rely on during detection: scene-motion coherence, inter-face appearance compatibility, interpersonal gaze alignment, and face-body consistency. Our findings provide empirical insights for multi-face deepfake detection depicting natural social interactions.

\item \textbf{Frame-level complete multi-face detection benchmark.} We emphasize the importance of detecting all fake faces in multi-face scenarios to enhance frame-level complete multi-face detection performance, which sets itself apart from existing single-face benchmarks.  We argue that detecting deepfakes in social settings is crucial and propose a human-centered paradigm to address this challenge. We hope this pioneering research will highlight the importance of deepfake detection in social contexts and the vital role of understanding human cognition in combating this growing threat.


\end{itemize}

\section{Related Work}
\label{sec:formatting}

\textbf{Multi-Face Deepfake Generation.} Understanding multi-face deepfake generation is essential for effective detection. Recent advancements include OpenForensics \cite{le2021openforensics}, FFIW \cite{zhou2021face}, ManualFake \cite{haiwei2022exploring}, and DF-Platter \cite{narayan2023df}. OpenForensics assesses manipulation feasibility using GAN-based synthesis \cite{pidhorskyi2020adversarial,shen2020interpreting} and Poisson blending. FFIW automates multi-face swapping with tools like DFL \cite{perov2020deepfacelab} and FSGAN \cite{nirkin2019fsgan}. ManualFake uses commercial software for synthesis and involves DFL, FSGAN, and Simswap \cite{chen2020simswap} manipulations, while DF-Platter generates a large-scale multi-face deepfake dataset using FSGAN \cite{nirkin2019fsgan} and FaceShifter \cite{li2019faceshifter}. However, these methods often overlook the social context of the faces, causing inconsistencies. Our human study highlights the importance of social context, which we incorporate into our multi-face deepfake detection framework.


\noindent \textbf{Multi-Face Deepfake Detection. }Most deepfake detection methods focus on single-face scenarios and fall into implicit clue-based \cite{meso,twobranch,fttwostream,recon,pellicer2024pudd,xu2023tall}, signal clue-based \cite{thinking,spsl,fwa,fakecatcher,ba2024exposing}, and semantic clue-based approaches \cite{emotions,lips,wang2023noise,consis,fttwostream,tan2023deepfake,choi2024exploiting}. These methods struggle in multi-face scenarios by ignoring contextual information and face relationships.

A few recent research has focused on multi-face deepfake detection. Limited works in this area include S-MIL \cite{li2020sharp}, Zhou et al. \cite{zhou2021face}, Ma et al. \cite{ma2022accurate}, FILTER \cite{lin2023exploiting},  COMISC \cite{zhang2024comics}, and MoNFAP \cite{miao2024mixture}. S-MIL  employs sharp multiple instance learning for video-level multi-face detection, while Zhou et al. use a discriminative attention model for the same purpose. COMISC  utilizes bi-grained contrastive learning, and MoNFAP uses noise extracts for detection, and FILTER  focuses on extracting facial aggregation features, and  Ma et al. use a VGG network to detect fake frames. 

While these methods achieve promising performance, they typically rely on black-box classifiers or individual heuristics, and often lack evaluation of frame-level complete multi-face detection. Our work leverages human cognitive insights from multiple observers to significantly improve frame-level complete multi-face detection performance, enabling a more effective defense against deepfake threats.

\noindent \textbf{Human Sensitivity in Face Perception.}
Neuroscience research has found that humans have dedicated neurobiological mechanisms for face recognition, primarily in the fusiform face area (FFA) and superior temporal sulcus \cite{Kanwisher1997, Haxby2000}. Human sensitivity in face perception plays a crucial role in social interactions and deception detection. Studies have shown that humans can rapidly recognize faces and detect subtle anomalies, such as unnatural textures or inconsistencies in expressions, which are often associated with deepfake or manipulated images \cite{farid2022creating, joslin2024double}. However, sensitivity to fake or abnormal faces varies based on context, prior exposure, and individual cognitive biases \cite{fan2014human,Nightingale2022}. While humans exhibit a general ability to detect manipulated faces, well-crafted deepfakes can still bypass perceptual defenses, leading to misjudgment \cite{Matern2019}. Our work is motivated by the above. We believe that understanding human perceptual mechanisms is essential for improving AI-driven deepfake detection and designing more effective countermeasures.

\noindent \textbf{Context-Aware Modeling of Human Groups. }
In addition to their innate sensitivity to faces, another characteristic of humans is their tendency to interact in groups. Research in this area focuses on understanding various group attributes, including activities, age, and gender.
In group activity recognition, researchers develop dynamic inference networks to analyze relationships among individuals. Notable advancements include graph network \cite{yuan2021spatio,xie2024active}, Dual-path Actor Interaction framework with Multi-scale Actor Contrastive Loss \cite{gallagher2009understanding}, and methods aligning local and global spatio-temporal views \cite{chappa2023spartan}. 
For age and gender detection in groups, contextual features prove beneficial. Gallagher et al. demonstrate that these features enhance age and gender prediction \cite{gallagher2009understanding}, while Rodriguez et al. introduce a feedforward attention mechanism to improve age recognition in group images \cite{rodriguez2017age}. 

Inspired by these studies, we highlight the importance of social contextual cues within groups for detection. We examine how humans detect these cues and integrate them into our detection framework.
\section{How Humans Detect Multi-Face Deepfake}
\begin{figure}[!t]
\centering
\includegraphics[width=0.8\columnwidth]{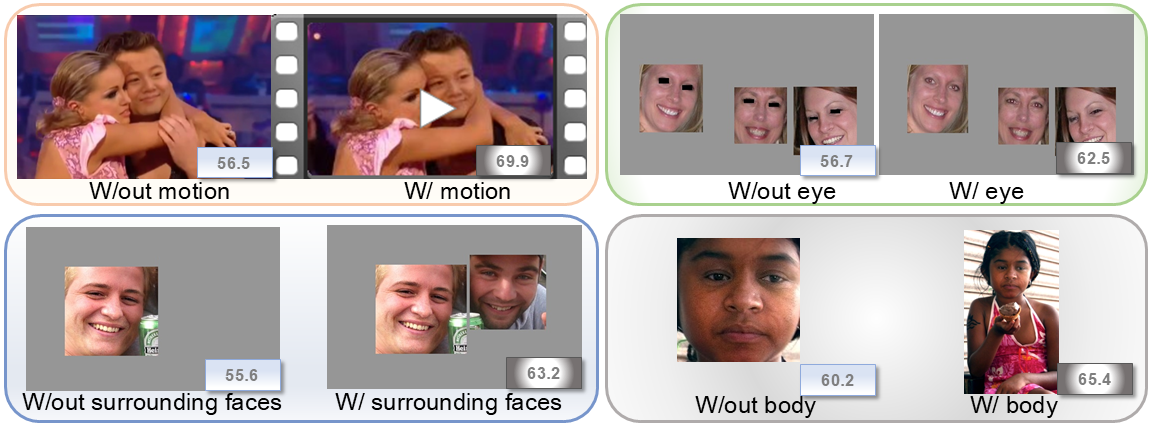} 
\caption{Examples of human studies with and without the four contextual features. The boxes display human performance in frame-level multi-face detection accuracy (\%).}
\label{fighumanablation}
\end{figure}

\subsection{Research Hypotheses}

Inspired by prior research on face perception and social scene understanding, we believe that incorporating human cognitive characteristics can enhance deepfake detection models, which motivates our human study. In this subsection, we outline our research hypotheses for the human study and provide the rationale behind each.

Firstly, face replacement in multi-face deepfake videos often introduces scene-motion inconsistencies, disrupting the natural scene arrangement and motion coherence among individuals \cite{agarwal2019limits, tian2024illumination}. Such inconsistencies appear as unnatural movements, jitter, or misalignments between faces and their surrounding context. Since humans naturally rely on scene coherence and motion smoothness to interpret group interactions \cite{ibrahim2016hierarchical, shehnepoor2022spatio}, we propose our first hypothesis:

\noindent\textbf{H1}: Deepfake techniques introduce scene-motion incoherence, which humans can identify as a key factor in deepfake detection.

Secondly, even with post-processing, deepfake faces often exhibit mismatches among faces, blending artifacts, or illumination inconsistencies within multi-face scenarios \cite{dolhansky2020deepfake,zhang2024comics}. Such discrepancies can create an unnatural appearance when a deepfaked face is compared to authentic faces in the same scene. Crowd analysis studies also find that inter-face appearance features are fundamental for understanding groups of people \cite{sharma2023scale,wang2021pixel}. Therefore, we propose our second hypothesis:

\noindent\textbf{H2}:  Deepfake faces exhibit inter-face appearance inconsistencies in resolution, color, or illumination in the scene, serving as contextual cues for human detection.

Thirdly, human gaze direction is a critical factor in both visual saliency and social perception. Research in gaze and psychology indicates that gaze plays an essential role in group settings \cite{hessels2020does, zhang2022gazeonce}. Studies have shown that gaze alignment is fundamental to social interactions, influencing attention and trustworthiness judgments \cite{langton2000eyes, sharma2024review}. Deepfake synthesis often fails to maintain natural gaze consistency, resulting in mismatches between the faked face and others in the scene \cite{liy2018exposingaicreated, cartella2024unveiling}. Building on these findings, we propose the third hypothesis for our human study: 

\noindent\textbf{H3}: Inconsistencies in gaze direction between deepfake faces and other individuals in a multi-face scene will be a detectable cue for humans. 

Lastly, deepfake generation methods often overlook body-face coherence, as most models focus primarily on facial synthesis rather than holistic body alignment \cite{zhou2017two, chakraborty2024role}. This lack of contextual awareness can lead to discrepancies between the generated face and body, particularly in terms of age and gender. Research on human behavior in groups has shown that age and gender are crucial factors for autonomous detection \cite{demarest2000body, wang2017gender}. Based on this, we propose the fourth hypothesis:

\noindent\textbf{H4}: Deepfake faces may show inconsistencies in body age and gender, providing an additional cue for detection.



\subsection{Human Study}

Based on our research hypotheses, we conduct a two-phase human study to explore human detection of multi-face deepfakes. In the first phase, we randomly selected $2,000$ multi-face deepfake videos and images from the OpenForensics \cite{le2021openforensics}, FFIW \cite{zhou2021face}, and DF-Platter \cite{narayan2023df}, with each video lasting approximately $20$ seconds. These datasets are the available benchmarks of current multi-face deepfakes. Four university students were recruited for this phase. Each participant was assigned $500$ videos and images to review. They were compensated at \$10 per hour. 
    
Participants documented the fake faces they identified, noted their reasons. They reviewed images and videos directly via a PC media player, without needing frame-by-frame analysis. Only identifications matching the dataset labels were considered valid. Participants categorized the  $500$ samples according to the detection cues, so that we can calculate the prevalence of each cue across all samples. 

The first phase identified $8$ primary indicators, summarized in Fig. \ref{fig1-challenge}. These were distilled into $4$ hypotheses for multi-face deepfake detection: scene-temporal artifacts (H1), inter-face appearance anomalies (H2), gaze direction inconsistencies (H3), and mismatches between body and face movements (H4). Using these findings, we designed the second phase of our study to explore how these cues influence detection accuracy. 

In the second phase, we sampled an additional 920 videos and images from our dataset pool and manipulated the fake faces across various scenarios to examine the impact of multiple contextual cues. These scenarios—blocked motions, blocked surrounding faces, blocked eyes, and blocked bodies—each isolates one of the four key contextual cues identified in the first phase (as shown in Fig. \ref{fighumanablation}). We recruited 20 participants from the online crowdsourcing platform Amazon Mechanical Turk \cite{crump2013evaluating} to identify the cues. To ensure reliability, participants are selected based on their approval rate and demographic suitability. Results in Fig. \ref{fighumanablation} demonstrate that performance improves when incorporating these contextual features, highlighting their significance in human detection.

\subsection{Human Cues in Detecting Deepfakes}

As shown in Fig. \ref{fig1-challenge}, four types of factors were identified to assist in human detection, with an emphasis on contextual elements. Minor cues (1.8\%) like background-text consistency were excluded from our integration. The most common cue, accounting for 34.2\%, stems from discontinuous motions between preceding and following frames, supporting H1. Additional significant factors, together representing 31.5\%, include inconsistencies that emerge in the inter-face context. These encompass variations in face resolution, mismatched lighting, color inconsistencies, and artifacts that appear in some faces but not others, supporting H2. The context of gaze within groups also plays a crucial role, with abnormal gaze direction relative to the camera accounting for 25.0\% of the cases (H3). Lastly, within the context of body and face alignment, discrepancies in age and gender between faces and bodies contribute 7.5\% to detection difficulties (H4).


\section{Context-Aware Multi-Face Detection}
\begin{figure*}[t]
\centering
\includegraphics[width=1.8\columnwidth]{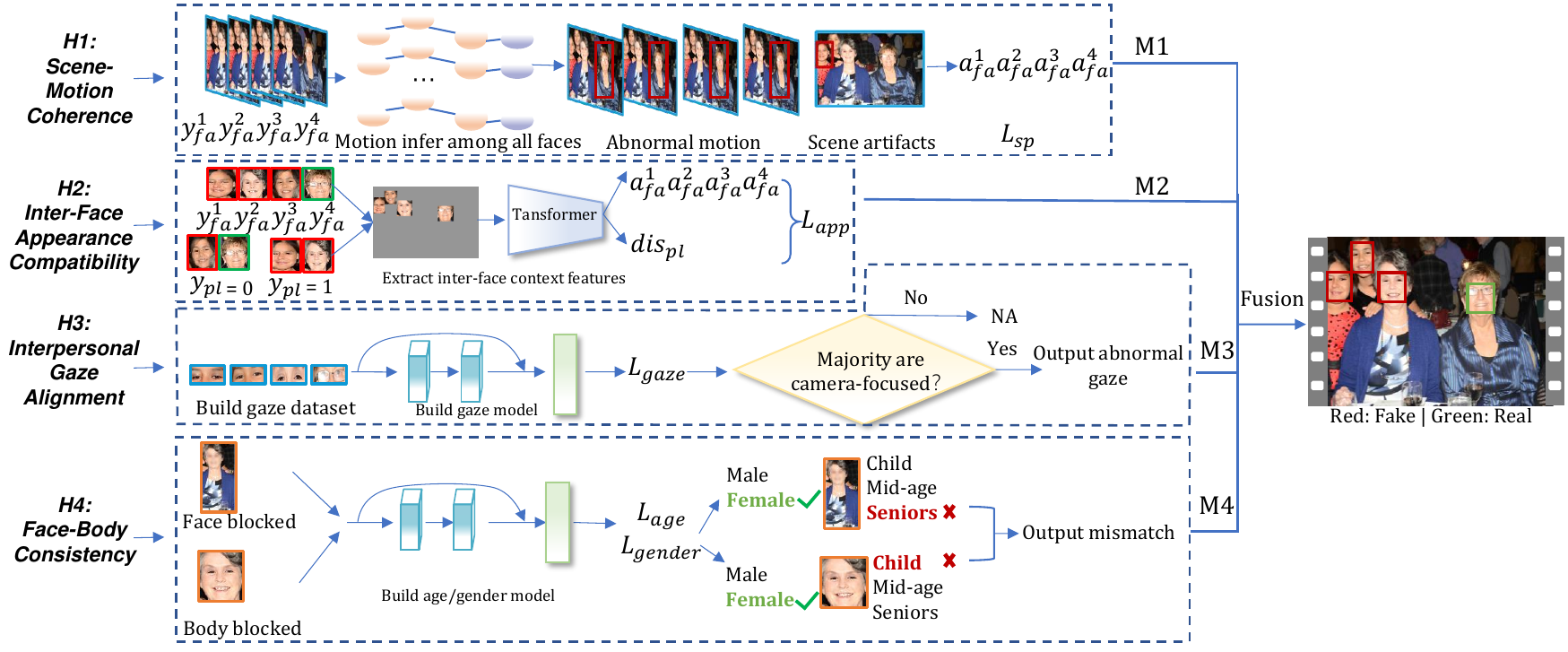} 
\caption{\textsf{HICOM} leverages human-inspired cues ($H1-H4$) derived from human studies to detect all fake faces within multi-face settings.}


\label{figcontext}
\end{figure*}
\subsection{Overall Framework and Design Rationale}

Leveraging insights from our human study, we propose a context-aware multi-face deepfake detection method that integrates human-derived reasons. Unlike methods that merely superimpose models,  each module of \textsf{HICOM} is inspired by specific human-reported cues. According to the H1 (scene-motion coherence), we develop a module that integrates facial and contextual features from preceding and following frames to expose unnatural motion and scene inconsistencies. We address the H2 (inter-face appearance compatibility) by developing an inter-face appearance module. This module enhances detection performance by combining single-face classification with multi-face comparisons. Inspired by H3 (interpersonal gaze alignment), we devise a gaze module that isolates eye regions to model gaze behavior, identifying anomalies such as a single eye not aligning with others directed toward the camera. Finally, we utilize H4 (face-body consistency) to devise a body-face module that independently assesses the age and gender of the face and body, detecting inconsistencies between them.

As illustrated in Fig. \ref{figcontext}, \textsf{HICOM} comprises four modules (M1–M4), each inspired by specific human-cognitive insights.  Rather than merely superimposing models, \textsf{HICOM} is grounded in human cognitive studies, with each module designed to integrate specific human-inspired contextual features. We leverage specialized feature extraction strategies, including inference from \cite{yuan2021spatio}, Transformer networks \cite{vaswani2017attention}, and ResNet architectures \cite{resnet}, tailored specifically for each module. Crucially, this modular design ensures robustness: even if three modules fail to detect anomalies, the remaining module can independently identify deepfake cues.

\subsection{Scene-Motion Module}
H1 identifies scene-motion inconsistency as the critical cue for deepfake detection, prompting us to develop a module that simulates this perception. The scene-motion module constructs multi-scale features by extracting detailed information from each face and its surrounding regions. By inferring motions and extracting scene inconsistencies across facial and background context over time, this module effectively detects scene-motion inconsistencies in fake faces. 

\noindent \textbf{Multi-Scale Feature Extraction. } We extract multi-scale features from sequences of images. The input data is reshaped to capture both scene and motion dimensions, allowing the network to generate comprehensive multi-scale feature representations across multiple frames. These features are refined using RoIAlign \cite{he2017mask}, which focuses on specific regions corresponding to detected faces and backgrounds. By embedding the extracted features through a fully connected layer, the model enhances its ability to detect scene-motion inconsistencies in multi-face scenarios.

\noindent \textbf{Scene-Motion Inference. } The scene-motion module infers motions and detects inconsistencies by analyzing each face and scene features across time. Inference network \cite{yuan2021spatio} is then used to focus on significant motion patterns, refining the detection of scene-motion inconsistencies. By combining features across time, the model generates a comprehensive representation that enhances detection accuracy.

The output features are processed to predict face-level and frame-level complete multi-face detection scores, allowing the model to identify deepfake manipulations effectively. Cross-entropy loss optimizes the model during training, ensuring robust detection across multiple faces.\small\begin{equation}
\mathcal{L}_{sp} = \lambda_{\text{fa}}  \text{CE}({a}_{fa}, y_{fa}) +  \lambda_{\text{fr}}\text{CE}({a}_{fr}, y_{fr}),
\end{equation}
where $\text{CE}({a}_{fa}, y_{fa})$ represents the face-level cross entropy loss for faces ${a}_{fa}$ with the true label $y_{fa}$, and $\lambda_{\text{fa}}$ and $\lambda_{\text{fr}}$ represents the equal weight for the face-level loss and frame-level loss, and $\text{CE}({a}_{fr}, y_{fr})$ represents the frame-level cross entropy loss for frames ${a}_{fr}$ with the true label $y_{fr}$, which checks if all faces in the frame are correctly predicted.
\subsection{Inter-Face Appearance  Module}
H2 underscores the importance of inter-face context in deepfake detection, prompting the design of an inter-face appearance module. The inter-face appearance  module focuses on face regions and the comparisons of multi-faces in a frame. By extracting facial features and comparing different faces, inter-face appearance module can detect the inconsistency among faces.

\noindent \textbf{Inter-Face Comparisons. } We crop face regions and use the aforementioned Transformer for model building. To compare multi-face features, we combine a contrastive loss and cross entropy loss for training.
\small\begin{align}
\mathcal{L}_{app} &=  \textstyle \text{CE}({a}_{fa}, y_{fa})+\textstyle \frac{\lambda_{comp}}{N_{comp}} \sum\nolimits_{j=1}^{N_{comp}}  \left[ y_{pl}^j \cdot {dis}_{pl}^j \right. \notag \\
& \quad \left. + (1-y_{pl}^j ) \cdot \max(0, \text{margin} - {dis}_{pl}^j) \right],
\end{align}
where \(\mathcal{L}_{app}\) denotes the total combined loss, $N_{comp}$ denotes the number of pair of samples, $y_{pl}^j$  denotes the binary label for the \(j\)-th pair of faces (where $y_{pl}^j=1$ if the faces are with similar label and $y_{pl}^j=0$ if they are dissimilar), ${dis}_{pl}^j$ denotes the Euclidean distance between the feature vectors of pair of faces, \(\text{margin}\)  denotes the minimum distance required for dissimilar pairs and is set to a default value of $1.0$,  {$\lambda_{comp}$} is empirically setted as $0.3$. 
\subsection{Gaze Module}
Our human study and previous work \cite{kulkarni2020can,MUGGLE} show that outlier observers often have gaze points that do not align with the group's common gaze, yet these outliers are not necessarily fake. To reduce false positives, the gaze module identifies abnormal gazes by analyzing eye regions to determine if the gaze is locked on the camera. According to H3, we design a gaze module that filters out multi-face videos and images where most faces are not looking at the camera. It then detects abnormal gazes by checking for camera-focused gazes. Building a trained gaze-locking model, we apply a decision strategy to identify abnormal gazes.

\noindent \textbf{Gaze Locking Model Construction.} We crop eye regions from sequences of images or frames and label them based on whether the gaze is directed at the camera. To expand our dataset, we use the Columbia Gaze-DataSet \cite{smith2013gaze}, which includes data with diverse head poses and gaze directions, providing robust data for training. We then pretrain a model using the Columbia Gaze-DataSet and a Resnet for gaze classification. Thereafter, we use our built dataset to train the Resnet. The model is optimized with cross-entropy loss ($\mathcal{L}_{gaze}$) calculated from ground-truth and predicted gaze labels and is saved when validation loss converges.

\noindent \textbf{Gaze Abnormal Detection. }  Not all faces with outlier gazes are fake \cite{chen2019unsupervised,ghosh2023automatic}. The module disregards multi-face images and videos where most faces are not looking at the camera. A face is flagged as fake only if most faces in the frame are looking at the camera while a few outliers are not. This is defined as:
\small\begin{equation}
a_i\!=\!
\begin{cases}
NA,&\text{if majority faces not looking at camera},\\
1, &\text{if } a_i  \text{ is off-camera and } (n_{L} \!-\! n_{O}\!>\! 1 \text{ or } n_{T} \!=\! 2),\\
0, &\text{if } a_i \text{ is not off-camera}.
\end{cases}
\end{equation}
where $n_{L}$, $n_{O}$, and $n_{T}$ represent the number of faces looking at the camera,  not looking at the camera, the total number of faces in the image or frame.

\subsection{Body-Face Module}
H4 underscores the importance of body-face context in deepfake detection, motivating the design of a dedicated body-face module. This module detects mismatches between the face and body in terms of age and gender.



\noindent \textbf{Face Block and Body Block.}
Body-blocked regions emphasize facial appearance, while face-blocked regions highlight clothing and posture features. To isolate body features for age and gender modeling, we apply GaussianBlur \cite{flusser2015recognition} to block the face areas within each body region. For face-only modeling, we crop faces from the images or frames to block body regions effectively.

\noindent \textbf{Age \& Gender Model Construction.}
For age model training, we categorize cropped faces and preprocessed body images into three groups: child, middle-aged, and senior. For gender model training, we classify them as male or female. Using the IMDB-WIKI dataset \cite{rothe2015dex}, we train Resnet to obtain a trained age and gender model, ensuring high age/gender detection performance. We then use this pretrained model to extract age and gender features.

 \noindent \textbf{Mismatch Detection. }We optimize the age and gender models with cross-entropy losses ($\mathcal{L}_{age}$ and $\mathcal{L}_{gender}$) and save the models upon convergence. Let ${ag}_i^{body}$ denote the predicted age or gender for the $i$-th body corresponding to the face, and ${ag}_i^{face}$ denote the predicted age or gender for the $i$-th face corresponding to the body. Detection is determined by:
\small\begin{equation}
 a_i = 
\begin{cases}
1, & \text{if }  ({ag}_i^{face})\neq ({ag}_i^{body}), \\
0, & \text{otherwise}.
\end{cases}
\end{equation}
\noindent \textbf{Effects of the Module. }Not all multi-face video images exhibit detectable body-face mismatches in age and gender. Therefore, this module acts as an auxiliary to other modules. When other modules fail to detect all fake faces in multi-face images, this module’s results can supplement them, improving frame-level complete multi-face detection performance.
\subsection{Module Combination}
Since our human study shows that cues from {M1} and {M2} are more significant than those of {M3} and {M4} for detection, M1 and M2 serve as the primary components in our framework. In contrast,  M3 and M4 are designed to provide complementary support.  When M1 and M2 miss a fake face, M3 and M4 help identify these inconsistencies, thereby enhancing frame-level complete multi-face detection performance. 
Inspired by these insights and previous literature \cite{anwar2023image,mishra2022hardly}, we fuse the outputs of these modules using an XOR operation, ensuring that any detected anomaly leads to a fake face prediction. The effectiveness of this fusion strategy is discussed further in the Supplementary Material.

\section{Evaluation of Multi-Face Detection}

\subsection{Experimental Settings}
\begin{table*}[htbp]
\small
\begin{center}

\renewcommand{\tabcolsep}{2mm} 
\renewcommand{\arraystretch}{1.0}
\begin{tabular}{ccccccccccccc}

\noalign{\hrule height 0.7pt}

\multirow{2}{*}{Method}&\multicolumn{4}{c} {FFIW}&\multicolumn{4}{c} {OpenForencics}& \multicolumn{4}{c} {DF-Platter}\\ 

    \cmidrule(r){2-3} \cmidrule(r){4-5}\cmidrule(r){6-7}\cmidrule(r){8-9} \cmidrule(r){10-11}\cmidrule(r){12-13}&FAC  &FAU &FCAC &FCAU&FAC  &FAU &FCAC &FCAU&FAC  &FAU &FCAC &FCAU\\
     
              \noalign{\hrule height 0.7pt}


{{SBI}$^{*}$ \cite{shiohara2022detecting}}&$94.0$&$94.2$&$84.1$&$85.6$& $\phantom{^{\dag}}92.8^{\dag}$&$\phantom{^{\dag}}98.8^{\dag}$& $83.4$&$85.7$ & $95.7$&$96.3$& $88.7$&$88.9$\\
{{TALL}$^{*}$ \cite{wang2023noise}}&$94.6$&$95.5$&$88.9$&$89.7$& $98.2$&$98.4$ &$93.1$&$94.6$&$96.8$&$96.9$& $90.2$&$91.7$ \\

{{Li et al.}$^{*}$ \cite{lin2024preserving}}&$86.3$&$91.1$&$77.2$&$78.9$& $91.1$&$93.4$& $80.7$&$81.9$ & $93.9$&$95.0$& $89.2$&$89.5$\\
\hline
 {Zhou et al.} \cite{zhou2021face}&$85.4$&$85.9$&$72.3$&$73.6$&$93.2$&$94.8$&$86.5$&$87.8$& $90.4$&$91.6$& $80.4$&$80.6$\\
{{Ma et al.} \cite{ma2022accurate}}&$88.4$&$91.5$&$82.5$&$83.2$&$96.4$&$98.5$&$87.6$&$88.2$& $95.2$&$96.5$& $89.2$&$89.9$\\
{{FILTER} \cite{lin2023exploiting}}&$92.5$&$94.4$&$84.9$&$85.4$&$\phantom{^{\dag}}99.0^{\dag}$&$\phantom{^{\dag}}\textbf{99.9}^{\dag}$&$93.6$&$93.7$& $96.8$&$97.5$& $89.5$&$90.6$\\
{{MoNFAP} \cite{miao2024mixture}}&$91.7$&$94.3$&${80.2}$&$82.1$& $99.1$&$\phantom{^{\dag}}\textbf{99.9}^{\dag}$& $89.6$&$92.3$& $92.6$&$93.7$& $88.4$&$89.3$\\
{{COMISC} \cite{zhang2024comics}}&$93.2$&$94.7$&${85.0}$&$85.6$& $98.4$&$99.5$& $93.7$&$94.8$& $93.4$&$94.8$& $89.2$&$89.7$\\
{\textsf{HICOM}}
&${\textbf{94.7}}$&$\textbf{{95.9}}$&$\textbf{{91.3}}$&${\textbf{92.1}}$&$\textbf{{99.3}}$&$\textbf{99.9}$&$\textbf{{97.8}}$&$\textbf{98.9}$&$\textbf{{97.2}}$&$\textbf{98.4}$&$\textbf{{93.5}}$&$\textbf{94.6}$\\

\noalign{\hrule height 0.7pt}

\end{tabular}
\end{center}

\caption{Comparisons of in-dataset detection performance  between \textsf{HICOM} and other methods on multi-face
datasets. For all tables, results marked with {\dag} are cited from FILTER \cite{lin2023exploiting}. Single-face methods are denoted by {*}, while multi-face methods are unmarked.}
\label{indatasets}
\vspace{-0.2cm}
\end{table*}
\textbf{Datasets.} We conduct experiments using four benchmark multi-face deepfake datasets: FFIW \cite{zhou2021face}, OpenForensics \cite{le2021openforensics}, DF-Platter \cite{narayan2023df}, and ManualFake \cite{haiwei2022exploring}, which are all widely-used benchmark datasets on multi-face deepfake. FFIW is a real-world multi-face deepfake video dataset, with frames containing up to $15$ faces. OpenForensics comprises GAN-generated images with an average of $2.9$ faces per image. Since OpenForensics is image-based, we replicate each image to create a sequence for input into the M1. DF-Platter is a multi-face deepfake video dataset with 2-5 faces per video. ManualFake provides multi-face deepfake versions transmitted through online social networks. Following MoNFAP \cite{miao2024mixture}, we use ManualFake to evaluate generalization in untrained real-world scenarios.

\noindent\textbf{Implementation Details.} We use the Adam optimizer with an initial learning rate of $1 \times 10^{-4}$, training for 120 epochs, and applying a decay rate of $1/3$ every 10 epochs. The size of M1 is $720 \times 1280$, while other modules use a size of $224 \times 224$. Experiments are conducted on NVIDIA H100 80GB GPUs.

\noindent\textbf{Metrics.} We report face-level ACC (FAC), face-level AUC (FAU), frame-level complete multi-face detection ACC (FCAC), and frame-level complete multi-face detection AUC (FCAU) scores. Face-level metrics assess each face independently, while frame-level complete multi-face detection metrics evaluate the detection of each face within that frame.

\noindent\textbf{Baselines.} We compare \textsf{HICOM} with representative single-face detection methods:  SBI \cite{shiohara2022detecting},  TALL \cite{xu2023tall}, and Li et al. \cite{lin2024preserving}, as well as the limited number of recently published SOTA multi-face detection methods, including Zhou et al. \cite{zhou2021face}, Ma et al. \cite{ma2022accurate}, FILTER \cite{lin2023exploiting}, MoNFAP \cite{miao2024mixture}, and COMISC \cite{zhang2024comics}.
\subsection{In-Dataset Detection Performance.}
We conduct in-dataset experiments on FFIW, OpenForensics, and DF-Platter, using the same datasets for both training and testing. As shown in Table \ref{indatasets}, while both single-face and multi-face detection methods perform well in face-level metrics, they degrade in frame-level complete multi-face detection metrics. However, \textsf{HICOM} achieves average improvements of 3.3\% in FCAC, and 3.1\% in FCAU compared to the next best results. This success stems from the method's thorough consideration of contextual features, including scene-motion coherence, inter-face appearance compatibility, interpersonal gaze alignment, and face-body consistency in terms of age and gender, ensuring comprehensive detection and minimizing missed fakes.

\begin{table}[!t]
\small
\begin{center}

\renewcommand{\tabcolsep}{2.3mm} 
\renewcommand{\arraystretch}{1.0}
\begin{tabular}{ccccc}
\noalign{\hrule height 0.7pt}

\multirow{2}{*}{Method}&\multicolumn{4}{c} {OpenForensics with Perturbations}\\


  \cmidrule(r){2-5} &FAC  &FAU &FCAC &FCAU\\
     
              \noalign{\hrule height 0.7pt}


{{SBI}$^{*}$ \cite{shiohara2022detecting}}&$\phantom{^{\dag}}74.7^{\dag}$&$\phantom{^{\dag}}82.5^{\dag}$&$66.1$&$67.4$\\
{{TALL}$^{*}$ \cite{wang2023noise}}&$90.7$&$96.5$&$77.1$&$78.4$ \\

{{Li et al.$^{*}$ }\cite{lin2024preserving}}&$75.6$&$75.8$&$63.7$&$64.9$ \\
\hline
{{Zhou et al. }\cite{zhou2021face}}&$78.9$&$79.5$&$64.7$&$68.9$\\
{{Ma et al. }\cite{ma2022accurate}}&$78.4$&$81.6$&$63.6$&$63.9$\\
{{FILTER }\cite{lin2023exploiting}}&$\phantom{^{\dag}}89.0^{\dag}$&$\phantom{^{\dag}}96.9^{\dag}$&$74.3$&$76.8$\\
{{MoNFAP} \cite{miao2024mixture}}&$87.3$&$89.2$&${72.8}$&$74.9$\\
{{COMISC} \cite{zhang2024comics}}&$88.2$&$92.1$&${73.0}$&$74.5$\\
{\textsf{HICOM}}
&{\textbf{91.2}}&\textbf{{97.5}}&\textbf{{78.6}}&\textbf{{81.2}}\\

\noalign{\hrule height 0.7pt}

\end{tabular}
\end{center}
\vspace{-0.3cm}
\caption{Robustness comparisons in unseen perturbations. }
\vspace{-0.3cm}
\label{robust}

\end{table}
\subsection{Model Generalizaiblity}
\noindent\textbf{Robustness to Unseen Real-World Perturbations. }Real-life deepfakes often involve various perturbations, and OpenForensics simulates this by providing six types: color manipulation, edge manipulation, block-wise distortion, image corruption, convolution mask transformation, and external effects. To assess \textsf{HICOM}'s robustness to these unseen perturbations, we conduct experiments on OpenForensics, where none of the perturbations were included in the training process. Results in Table \ref{robust} show that while existing methods struggle to generalize to unseen perturbations, \textsf{HICOM} achieves an average improvement of 1.5\% FCAC and 2.8\% FCAU over the previous best results. This improvement stems from \textsf{HICOM}'s reliance on contextual features, which are less dependent on specific training data and more resilient to perturbations. For example, abnormal gaze and mismatches in age and gender between faces and bodies, identified during training, remain detectable even under perturbations in the test set. Additional experiments on videos with unknown compression factors are detailed in the Supplementary Material.

\begin{table}[!t]
\small
\begin{center}

\renewcommand{\tabcolsep}{0.25mm} 
\renewcommand{\arraystretch}{1.0}
\begin{tabular}{ccccccccc}
\noalign{\hrule height 0.7pt}

\multirow{3}{*}{Method}&\multicolumn{4}{c} {DF-Platter to ManualFake}&\multicolumn{4}{c} {FFIW to ManualFake}\\ 


    \cmidrule(r){2-5} \cmidrule(r){6-9} &FAC  &FAU &FCAC &FCAU&FAC  &FAU &FCAC &FCAU\\
     
              \noalign{\hrule height 0.7pt}



{{SBI}$^{*}$\cite{shiohara2022detecting}}&$69.1$&$70.7$&$59.3$&$60.9$&$70.3$&$71.4$&$63.3$&$63.9$\\
{{TALL}$^{*}$\cite{wang2023noise}}&$69.3$&$70.4$&$59.9$&$60.6$&$71.3$&$72.2$&$65.8$&$66.7$\\

{{Li et al.}$^{*}$ \cite{lin2024preserving}} &$68.3$&$69.4$&$56.6$&$57.5$&$69.7$&$69.9$&$58.1$&$59.3$\\
\hline
{{Zhou et al.} \cite{zhou2021face}}&$64.2$&$65.9$&$56.1$&$56.3$&$68.4$&$69.1$&$56.2$&$57.7$\\
{{Ma et al.} \cite{ma2022accurate}}&$63.8$&$64.7$&$55.1$&$56.2$&$68.3$&$69.6$&$56.3$&$56.9$\\
{{FILTER} \cite{lin2023exploiting}}&$68.1$&$69.4$&$56.8$&$57.7$&$66.4$&$68.3$&$61.8$&$62.2$\\
{{MoNFAP} \cite{miao2024mixture}}&$60.7$&$67.9$&$53.2$&$54.3$&$61.6$&$62.2$&$55.7$&$56.2$\\
{{COMISC} \cite{zhang2024comics}}&$67.7$&$68.9$&$58.2$&$59.4$&$68.8$&$69.9$&$62.6$&$63.3$\\
{\textsf{HICOM}}
&{\textbf{70.7}}&\textbf{{71.4}}&\textbf{{66.3}}&\textbf{{67.7}}&{\textbf{72.8}}&{\textbf{73.3}}&\textbf{{70.9}}&\textbf{{71.6}}\\
\noalign{\hrule height 0.7pt}
\end{tabular}
\end{center}
\vspace{-0.2cm}
\caption{Generalization comparisons in untrained datasets.}
\vspace{-0.2cm}
\label{crossdata}
\end{table}
\begin{table}[!t]
\small
\begin{center}

\renewcommand{\tabcolsep}{0.5mm} 
\renewcommand{\arraystretch}{1.0}
\begin{tabular}{ccccccc}
\noalign{\hrule height 0.7pt}

\multirow{2}{*}{Module}&\multicolumn{2}{c} {FFIW}&\multicolumn{2}{c} {OpenForencics}& \multicolumn{2}{c} {DF-Platter}\\ \cmidrule(r){2-3} \cmidrule(r){4-5}\cmidrule(r){6-7} &\multicolumn{1}{c} {FCAC}&\multicolumn{1}{c} {FCAU}&\multicolumn{1}{c} {FCAC}&\multicolumn{1}{c} {FCAU}&\multicolumn{1}{c} {FCAC}&\multicolumn{1}{c} {FCAU}\\

              \noalign{\hrule height 0.7pt}

{\textmd{M1}}&$87.7$&$89.0$&$93.9$&$95.6$&$90.4$&$91.1$\\
{\textmd{M1+M2}}&$89.9$&$90.6$&$95.7$&$97.3$&$92.0$&$92.8$\\
{\textmd{M1+M2+M3}}&$90.6$&$91.4$&$97.2$&$98.3$&$93.3$&$94.0$\\
{\textmd{M1+M2+M3+M4}}
&$\textbf{{91.3}}$&${\textbf{92.1}}$&$\textbf{{97.8}}$&$\textbf{98.9}$&$\textbf{{93.5}}$&$\textbf{94.6}$\\


\noalign{\hrule height 0.7pt}

\end{tabular}
\end{center}
\vspace{-0.2cm}
\caption{Ablation study - Comparisons of frame-level complete multi-face detection performance in different modules. }

\label{ablation}
\end{table}
\noindent\textbf{Generalization to Unseen Dataset. } To evaluate the generalization of the identified cues, we conduct cross-dataset experiments, training the model on DF-Platter or FFIW and testing it on ManualFake. Notably, ManualFake was not used for human studies or model training, and OpenForensics was excluded from training as it contains only images rather than videos. The results presented in Table \ref{crossdata} indicate that all state-of-the-art methods exhibit a significant drop in performance, highlighting the substantial challenge of multi-face deepfake detection. Nevertheless, \textsf{HICOM} achieves an average improvement of 5.8\% at the frame-level complete multi-face detection accuracy, demonstrating that the cues inspired by human studies and incorporated into our design exhibit a certain degree of generalization. 

\noindent\textbf{Single-Face Detection.} Our method can be adapted to single-face scenarios. Specifically, we modify M1 to extract only scene-motion features and M2 to focus solely on single-face features, removing inter-face dependencies. M3 is excluded as it is not applicable, while M4 remains unchanged.  We evaluate this adaptation on the FF++ \cite{ffdata} dataset, demonstrating competitive performance against SOTA methods in single-face detection. Due to space constraints, detailed results are provided in Supplementary Material.
\subsection{Analyses and Discussions} \noindent\textbf{Effects of Four Modules.}
We progressively evaluate modules M1 through M4, as shown in Table \ref{ablation}. M1 alone achieves acceptable performance, while adding subsequent modules consistently improves results. M2 significantly boosts accuracy by combining classification and contrastive loss across multiple faces. M3 provides modest gains by targeting gaze anomalies but is less effective when gazes are dispersed. M4 contributes the least, as it only activates for specific face-body inconsistencies. Nonetheless, M3 and M4 remain essential for detecting faces missed by earlier modules.

\begin{figure}[!t]
\centering
\includegraphics[width=0.9\columnwidth]{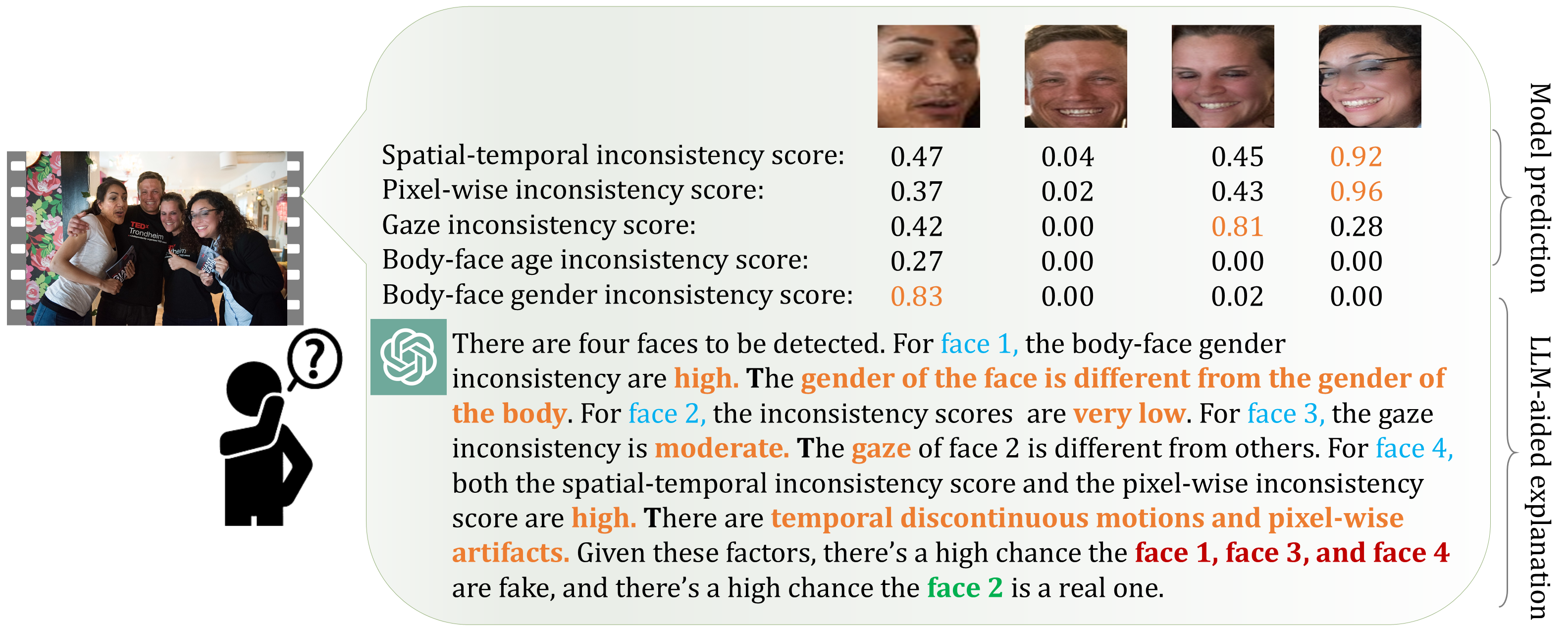} 
\caption{\textsf{HICOM} provides comprehensible explanations for its predictions through integration with an LLM.}
\label{figllm}
\vspace{-0.1cm}
\end{figure}
\begin{figure}[!t]
\centering
\includegraphics[width=0.5\columnwidth]{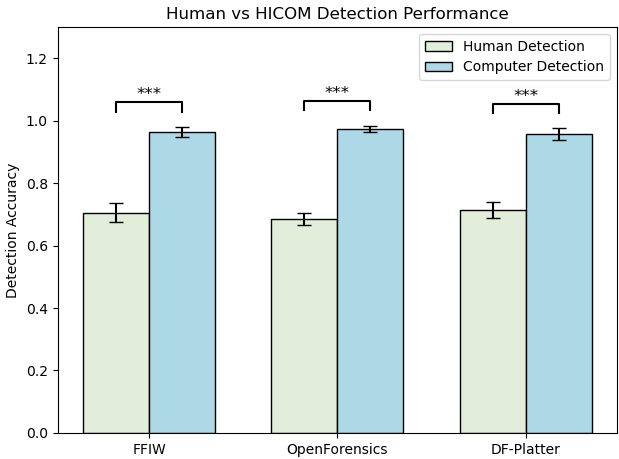} 
\caption{\textsf{HICOM} surpasses humans in multi-face detection.}
\label{fighuco}
\end{figure}

\noindent\textbf{LLM Explanation.} We use the output scores of M1, M2, M3, and M4 as prompts and invoke the ChatGPT API \cite{achiam2023gpt} to explain \textsf{HICOM}, which shows our model's potential to work with LLM to provide an explainable prediction.  
Results in Fig. \ref{figllm} show that each face is detected and explained as either fake or real, enhancing user trust and understanding of the detection.

\noindent\textbf{Human Detection Results.} We conduct human studies on multi-face detection using 300 randomly selected samples from FFIW, OpenForensics, and DF-Platter, comparing the results with \textsf{HICOM}. Each set is evaluated by $5$ AMT workers, with their averaged performance and standard error reported. We label a $p$-value less than 0.0005 with three stars, a $p$-value between 0.0005 and 0.005 with two stars, and a $p$-value greater than 0.005 with one star. Results in Fig. \ref{fighuco} show three stars for the $p$-value, indicating that \textsf{HICOM} outperforms human detection. This demonstrates its effectiveness in assisting users with multi-face deepfake detection.



\section{Conclusion}
This paper presents a novel framework that leverages crowdsourced human studies, approved by the university's Institutional Review Board (IRB), to systematically detect every single fake face in multi-face scenarios. Emphasizing cognitive processes as fundamental to deepfake detection, the proposed framework \textsf{HICOM} moves beyond traditional methods that rely solely on black-box classifiers or individual heuristics. Instead, it integrates human cognitive insights derived from multiple observers into the detection framework.  Quantitative results identify scene-motion coherence, inter-face appearance compatibility, interpersonal gaze alignment, and face-body consistency as key factors in multi-face deepfake detection. By incorporating these human-inspired cues, \textsf{HICOM} demonstrates how social context provides a richer context for distinguishing real from fake faces in group settings. This work represents a pioneering step in detecting multiple fake faces within social contexts. 

\noindent\textbf{Limitations. } Our findings are based on all benchmark multi-face deepfake datasets, with cues specifically designed for fake face detection. As a result, some cues (\emph{e.g.}, gaze alignment and face-body consistency) may not universally apply to natural images. Additionally, as deepfake techniques evolve, new contextual cues may emerge. However, our paradigm and hypotheses are grounded in fundamental human cognitive patterns, making them broadly applicable to future deepfakes.

While our modular approach offers flexibility and generalizability in complex multi-face scenarios, it is not optimized for end-to-end settings. Nevertheless, given the diverse manipulation cues across multiple faces, this design remains advantageous for accurate detection and interpretation. Moreover, it can be easily adapted to incorporate new cues as deepfake techniques evolve.



\noindent\textbf{Acknowledgements. }This work is supported by the Ministry of Education, Singapore, under its MOE AcRF TIER 3 Grant (MOE-MOET32022-0001).
{
    \small
    \bibliographystyle{ieeenat_fullname}
    \bibliography{main}

\begin{thebibliography}{85}
\providecommand{\natexlab}[1]{#1}
\providecommand{\url}[1]{\texttt{#1}}
\expandafter\ifx\csname urlstyle\endcsname\relax
  \providecommand{\doi}[1]{doi: #1}\else
  \providecommand{\doi}{doi: \begingroup \urlstyle{rm}\Url}\fi

\bibitem[Achiam et~al.(2023)Achiam, Adler, Agarwal, Ahmad, Akkaya, Aleman, Almeida, Altenschmidt, Altman, Anadkat, et~al.]{achiam2023gpt}
Josh Achiam, Steven Adler, Sandhini Agarwal, Lama Ahmad, Ilge Akkaya, Florencia~Leoni Aleman, Diogo Almeida, Janko Altenschmidt, Sam Altman, Shyamal Anadkat, et~al.
\newblock Gpt-4 technical report.
\newblock \emph{Open AI}, 2023.

\bibitem[Afchar et~al.(2018)Afchar, Nozick, Yamagishi, and Echizen]{meso}
Darius Afchar, Vincent Nozick, Junichi Yamagishi, and Isao Echizen.
\newblock Mesonet: a compact facial video forgery detection network.
\newblock In \emph{WIFS}, pages 1--7, 2018.

\bibitem[Agarwal and Varshney(2019)]{agarwal2019limits}
Sakshi Agarwal and Lav~R Varshney.
\newblock Limits of deepfake detection: A robust estimation viewpoint.
\newblock \emph{arXiv preprint arXiv:1905.03493}, 2019.

\bibitem[Anwar et~al.(2023)Anwar, Tahir, Fahad, and Kifayat]{anwar2023image}
Muhammad~Aqib Anwar, Syed~Fahad Tahir, Labiba~Gillani Fahad, and Kashif Kifayat.
\newblock Image forgery detection by transforming local descriptors into deep-derived features.
\newblock \emph{Applied Soft Computing}, 147:\penalty0 110730, 2023.

\bibitem[Ba et~al.(2024)Ba, Liu, Liu, Wu, Lin, Lu, and Ren]{ba2024exposing}
Zhongjie Ba, Qingyu Liu, Zhenguang Liu, Shuang Wu, Feng Lin, Li Lu, and Kui Ren.
\newblock Exposing the deception: Uncovering more forgery clues for deepfake detection.
\newblock In \emph{AAAI}, pages 719--728, 2024.

\bibitem[BBC(2025)]{Trumpbbc}
BBC.
\newblock Zelensky told to leave white house after angry spat with trump and vance.
\newblock \url{https://www.bbc.com/news/live/c625ex282zzt}, 2025.
\newblock Accessed: 2025-03-07.

\bibitem[Cao et~al.(2022)Cao, Ma, Yao, Chen, Ding, and Yang]{recon}
Junyi Cao, Chao Ma, Taiping Yao, Shen Chen, Shouhong Ding, and Xiaokang Yang.
\newblock End-to-end reconstruction-classification learning for face forgery detection.
\newblock In \emph{CVPR}, pages 4113--4122, 2022.

\bibitem[Cartella et~al.(2024)Cartella, Cuculo, Cornia, and Cucchiara]{cartella2024unveiling}
Giuseppe Cartella, Vittorio Cuculo, Marcella Cornia, and Rita Cucchiara.
\newblock Unveiling the truth: Exploring human gaze patterns in fake images.
\newblock \emph{IEEE Signal Processing Letters}, 2024.

\bibitem[Chakraborty and Naskar(2024)]{chakraborty2024role}
Rajat Chakraborty and Ruchira Naskar.
\newblock Role of human physiology and facial biomechanics towards building robust deepfake detectors: A comprehensive survey and analysis.
\newblock \emph{Computer Science Review}, 54:\penalty0 100677, 2024.

\bibitem[Chappa et~al.(2023)Chappa, Nguyen, Nelson, Seo, Li, Dobbs, and Luu]{chappa2023spartan}
Naga~VS Chappa, Pha Nguyen, Alexander~H Nelson, Han-Seok Seo, Xin Li, Page~Daniel Dobbs, and Khoa Luu.
\newblock Spartan: Self-supervised spatiotemporal transformers approach to group activity recognition.
\newblock In \emph{CVPR}, pages 5158--5168, 2023.

\bibitem[Chen et~al.(2020)Chen, Chen, Ni, and Ge]{chen2020simswap}
Renwang Chen, Xuanhong Chen, Bingbing Ni, and Yanhao Ge.
\newblock Simswap: An efficient framework for high fidelity face swapping.
\newblock In \emph{ACM MM}, pages 2003--2011, 2020.

\bibitem[Chen et~al.(2019)Chen, Deng, Pi, and Shi]{chen2019unsupervised}
Zhaokang Chen, Didan Deng, Jimin Pi, and Bertram~E Shi.
\newblock Unsupervised outlier detection in appearance-based gaze estimation.
\newblock In \emph{ICCVW}, 2019.

\bibitem[Choi et~al.(2024)Choi, Kim, Jeong, Baek, and Choi]{choi2024exploiting}
Jongwook Choi, Taehoon Kim, Yonghyun Jeong, Seungryul Baek, and Jongwon Choi.
\newblock Exploiting style latent flows for generalizing deepfake video detection.
\newblock In \emph{CVPR}, pages 1133--1143, 2024.

\bibitem[Ciftci et~al.(2020)Ciftci, Demir, and Yin]{fakecatcher}
Umur~Aybars Ciftci, Ilke Demir, and Lijun Yin.
\newblock Fakecatcher: Detection of synthetic portrait videos using biological signals.
\newblock \emph{TPAMI}, 2020.

\bibitem[Crump et~al.(2013)Crump, McDonnell, and Gureckis]{crump2013evaluating}
Matthew~JC Crump, John~V McDonnell, and Todd~M Gureckis.
\newblock Evaluating amazon's mechanical turk as a tool for experimental behavioral research.
\newblock \emph{PloS one}, 8\penalty0 (3):\penalty0 e57410, 2013.

\bibitem[Demarest and Allen(2000)]{demarest2000body}
Jack Demarest and Rita Allen.
\newblock Body image: Gender, ethnic, and age differences.
\newblock \emph{The Journal of Social Psychology}, 140\penalty0 (4):\penalty0 465--472, 2000.

\bibitem[Dolhansky et~al.(2020)Dolhansky, Bitton, Pflaum, Lu, Howes, Wang, and Ferrer]{dolhansky2020deepfake}
Brian Dolhansky, Joanna Bitton, Ben Pflaum, Jikuo Lu, Russ Howes, Menglin Wang, and Cristian~Canton Ferrer.
\newblock The deepfake detection challenge (dfdc) dataset.
\newblock \emph{arXiv preprint arXiv:2006.07397}, 2020.

\bibitem[Fan et~al.(2014)Fan, Wang, Ng, Tan, Herberg, and Koenig]{fan2014human}
Shaojing Fan, Rangding Wang, Tian-Tsong Ng, Cheston Y-C Tan, Jonathan~S Herberg, and Bryan~L Koenig.
\newblock Human perception of visual realism for photo and computer-generated face images.
\newblock \emph{ACM TAP}, 11\penalty0 (2):\penalty0 1--21, 2014.

\bibitem[Farah et~al.(1998)Farah, Wilson, Drain, and Tanaka]{farah1998special}
Martha~J Farah, Kevin~D Wilson, Maxwell Drain, and James~N Tanaka.
\newblock What is" special" about face perception?
\newblock \emph{Psychological review}, 105\penalty0 (3):\penalty0 482, 1998.

\bibitem[Farid(2022)]{farid2022creating}
Hany Farid.
\newblock Creating, using, misusing, and detecting deep fakes.
\newblock \emph{Journal of Online Trust and Safety}, 1\penalty0 (4), 2022.

\bibitem[Flusser et~al.(2015)Flusser, Farokhi, H{\"o}schl, Suk, Zitova, and Pedone]{flusser2015recognition}
Jan Flusser, Sajad Farokhi, Cyril H{\"o}schl, Tom{\'a}{\v{s}} Suk, Barbara Zitova, and Matteo Pedone.
\newblock Recognition of images degraded by gaussian blur.
\newblock \emph{TIP}, 25\penalty0 (2):\penalty0 790--806, 2015.

\bibitem[Gaither et~al.(2016)Gaither, Pauker, Slepian, and Sommers]{gaither2016social}
Sarah~E Gaither, Kristin Pauker, Michael~L Slepian, and Samuel~R Sommers.
\newblock Social belonging motivates categorization of racially ambiguous faces.
\newblock \emph{Social cognition}, 34\penalty0 (2):\penalty0 97--118, 2016.

\bibitem[Gallagher and Chen(2009)]{gallagher2009understanding}
Andrew~C Gallagher and Tsuhan Chen.
\newblock Understanding images of groups of people.
\newblock In \emph{CVPR}, pages 256--263, 2009.

\bibitem[Ghosh et~al.(2023)Ghosh, Dhall, Hayat, Knibbe, and Ji]{ghosh2023automatic}
Shreya Ghosh, Abhinav Dhall, Munawar Hayat, Jarrod Knibbe, and Qiang Ji.
\newblock Automatic gaze analysis: A survey of deep learning based approaches.
\newblock \emph{TPAMI}, 46\penalty0 (1):\penalty0 61--84, 2023.

\bibitem[Gu et~al.(2022)Gu, Chen, Yao, Ding, Li, and Ma]{consis}
Zhihao Gu, Yang Chen, Taiping Yao, Shouhong Ding, Jilin Li, and Lizhuang Ma.
\newblock Delving into the local: Dynamic inconsistency learning for deepfake video detection.
\newblock In \emph{AAAI}, pages 744--752, 2022.

\bibitem[Haiwei et~al.(2022)Haiwei, Jiantao, Shile, and Jinyu]{haiwei2022exploring}
Wu Haiwei, Zhou Jiantao, Zhang Shile, and Tian Jinyu.
\newblock Exploring spatial-temporal features for deepfake detection and localization.
\newblock \emph{arXiv preprint arXiv:2210.15872}, 2022.

\bibitem[Haliassos et~al.(2021)Haliassos, Vougioukas, Petridis, and Pantic]{lips}
Alexandros Haliassos, Konstantinos Vougioukas, Stavros Petridis, and Maja Pantic.
\newblock Lips don't lie: A generalisable and robust approach to face forgery detection.
\newblock In \emph{CVPR}, pages 5039--5049, 2021.

\bibitem[Haxby et~al.(2000)Haxby, Hoffman, and Gobbini]{Haxby2000}
James~V. Haxby, Elizabeth~A. Hoffman, and Maria~I. Gobbini.
\newblock The distributed human neural system for face perception.
\newblock \emph{Trends in Cognitive Sciences}, 4\penalty0 (6):\penalty0 223--233, 2000.

\bibitem[He et~al.(2016)He, Zhang, Ren, and Sun]{resnet}
Kaiming He, Xiangyu Zhang, Shaoqing Ren, and Jian Sun.
\newblock Deep residual learning for image recognition.
\newblock In \emph{CVPR}, pages 770--778, 2016.

\bibitem[He et~al.(2017)He, Gkioxari, Doll{\'a}r, and Girshick]{he2017mask}
Kaiming He, Georgia Gkioxari, Piotr Doll{\'a}r, and Ross Girshick.
\newblock Mask r-cnn.
\newblock In \emph{ICCV}, pages 2961--2969, 2017.

\bibitem[Hessels(2020)]{hessels2020does}
Roy~S Hessels.
\newblock How does gaze to faces support face-to-face interaction? a review and perspective.
\newblock \emph{Psychonomic Bulletin \& Review}, 27\penalty0 (5):\penalty0 856--881, 2020.

\bibitem[Hu et~al.(2021)Hu, Liao, Wang, and Qin]{fttwostream}
Juan Hu, Xin Liao, Wei Wang, and Zheng Qin.
\newblock Detecting compressed deepfake videos in social networks using frame-temporality two-stream convolutional network.
\newblock \emph{TCSVT}, 32\penalty0 (3):\penalty0 1089--1102, 2021.

\bibitem[Ibrahim et~al.(2016)Ibrahim, Muralidharan, Deng, Vahdat, and Mori]{ibrahim2016hierarchical}
Mostafa~S Ibrahim, Srikanth Muralidharan, Zhiwei Deng, Arash Vahdat, and Greg Mori.
\newblock A hierarchical deep temporal model for group activity recognition.
\newblock In \emph{CVPR}, pages 1971--1980, 2016.

\bibitem[Joslin et~al.(2024)Joslin, Wang, and Hao]{joslin2024double}
Matthew Joslin, Xian Wang, and Shuang Hao.
\newblock Double face: Leveraging user intelligence to characterize and recognize ai-synthesized faces.
\newblock In \emph{USENIX Security}, pages 1009--1026, 2024.

\bibitem[Kanwisher(2000)]{kanwisher2000domain}
Nancy Kanwisher.
\newblock Domain specificity in face perception.
\newblock \emph{Nature neuroscience}, 3\penalty0 (8):\penalty0 759--763, 2000.

\bibitem[Kanwisher et~al.(1997)Kanwisher, McDermott, and Chun]{Kanwisher1997}
Nancy Kanwisher, Josh McDermott, and Marvin~M. Chun.
\newblock The fusiform face area: A module in human extrastriate cortex specialized for face perception.
\newblock \emph{The Journal of Neuroscience}, 17\penalty0 (11):\penalty0 4302--4311, 1997.

\bibitem[Kulkarni et~al.(2020)Kulkarni, Patil, Parikh, Arora, and Atrey]{kulkarni2020can}
Omkar~N Kulkarni, Vikram Patil, Shivam~B Parikh, Shashank Arora, and Pradeep~K Atrey.
\newblock Can you all look here? towards determining gaze uniformity in group images.
\newblock In \emph{ISM}, pages 100--103, 2020.

\bibitem[Langton et~al.(2000)Langton, Watt, and Bruce]{langton2000eyes}
Stephen~RH Langton, Roger~J Watt, and Vicki Bruce.
\newblock Do the eyes have it? cues to the direction of social attention.
\newblock \emph{Trends in Cognitive Sciences}, 4\penalty0 (2):\penalty0 50--59, 2000.

\bibitem[Le et~al.(2021)Le, Nguyen, Yamagishi, and Echizen]{le2021openforensics}
Trung-Nghia Le, Huy~H Nguyen, Junichi Yamagishi, and Isao Echizen.
\newblock Openforensics: Large-scale challenging dataset for multi-face forgery detection and segmentation in-the-wild.
\newblock In \emph{ICCV}, pages 10117--10127, 2021.

\bibitem[Li et~al.(2019)Li, Bao, Yang, Chen, and Wen]{li2019faceshifter}
Lingzhi Li, Jianmin Bao, Hao Yang, Dong Chen, and Fang Wen.
\newblock Faceshifter: Towards high fidelity and occlusion aware face swapping.
\newblock \emph{arXiv preprint arXiv:1912.13457}, 2019.

\bibitem[Li et~al.(2020)Li, Lang, Chen, Mao, He, Wang, Xue, and Lu]{li2020sharp}
Xiaodan Li, Yining Lang, Yuefeng Chen, Xiaofeng Mao, Yuan He, Shuhui Wang, Hui Xue, and Quan Lu.
\newblock Sharp multiple instance learning for deepfake video detection.
\newblock In \emph{ACM MM}, pages 1864--1872, 2020.

\bibitem[Li and Lyu(2019)]{fwa}
Yuezun Li and Siwei Lyu.
\newblock Exposing {D}eepfake videos by detecting face warping artifacts.
\newblock In \emph{CVPRW}, pages 46--52, 2019.

\bibitem[Lin et~al.(2024{\natexlab{a}})Lin, Yi, Wang, Li, Jingyi, and Shen]{lin2023exploiting}
Chenhao Lin, Fangbin Yi, Hang Wang, Qian Li, Deng Jingyi, and Chao Shen.
\newblock Exploiting facial relationships and feature aggregation for multi-face forgery detection.
\newblock \emph{TIFS}, 19:\penalty0 8832--8844, 2024{\natexlab{a}}.

\bibitem[Lin et~al.(2024{\natexlab{b}})Lin, He, Ju, Wang, Ding, and Hu]{lin2024preserving}
Li Lin, Xinan He, Yan Ju, Xin Wang, Feng Ding, and Shu Hu.
\newblock Preserving fairness generalization in deepfake detection.
\newblock In \emph{CVPR}, pages 16815--16825, 2024{\natexlab{b}}.

\bibitem[Liu et~al.(2021)Liu, Li, Zhou, Chen, He, Xue, Zhang, and Yu]{spsl}
Honggu Liu, Xiaodan Li, Wenbo Zhou, Yuefeng Chen, Yuan He, Hui Xue, Weiming Zhang, and Nenghai Yu.
\newblock Spatial-phase shallow learning: rethinking face forgery detection in frequency domain.
\newblock In \emph{CVPR}, pages 772--781, 2021.

\bibitem[Liy and InIctuOculi(2018)]{liy2018exposingaicreated}
Chang~M Liy and LYUS InIctuOculi.
\newblock Exposingaicreated fakevideosbydetectingeyeblinking.
\newblock In \emph{WIFS}, 2018.

\bibitem[Ma and Liu(2022)]{ma2022accurate}
Zekun Ma and Bin Liu.
\newblock Accurate and time-saving deepfake detection in multi-face scenarios using combined features.
\newblock In \emph{ICCSSE}, pages 378--382, 2022.

\bibitem[Masi et~al.(2020)Masi, Killekar, Mascarenhas, Gurudatt, and AbdAlmageed]{twobranch}
Iacopo Masi, Aditya Killekar, Royston~Marian Mascarenhas, Shenoy~Pratik Gurudatt, and Wael AbdAlmageed.
\newblock Two-branch recurrent network for isolating deepfakes in videos.
\newblock In \emph{ECCV}, pages 667--684, 2020.

\bibitem[Matern et~al.(2019)Matern, Riess, and Stamminger]{Matern2019}
Florian Matern, Christian Riess, and Marc Stamminger.
\newblock Exploiting visual artifacts to expose deepfakes and face manipulations.
\newblock \emph{CVPRW}, pages 288--295, 2019.

\bibitem[Miao et~al.(2024)Miao, Chu, Gong, Tan, Jin, Zhuang, Luo, Hu, and Yu]{miao2024mixture}
Changtao Miao, Qi Chu, Tao Gong, Zhentao Tan, Zhenchao Jin, Wanyi Zhuang, Man Luo, Honggang Hu, and Nenghai Yu.
\newblock Mixture-of-noises enhanced forgery-aware predictor for multi-face manipulation detection and localization.
\newblock \emph{arXiv preprint arXiv:2408.02306}, 2024.

\bibitem[Mishra et~al.(2022)Mishra, Sinha, Mitra, and Sahoo]{mishra2022hardly}
Suchintan Mishra, Harshit~Raj Sinha, Tushar Mitra, and Manadeepa Sahoo.
\newblock I hardly lie: A multistage fake news detection system.
\newblock In \emph{Biologically Inspired Techniques in Many Criteria Decision Making: Proceedings of BITMDM 2021}, pages 253--261. Springer, 2022.

\bibitem[Mittal et~al.(2020)Mittal, Bhattacharya, Chandra, Bera, and Manocha]{emotions}
Trisha Mittal, Uttaran Bhattacharya, Rohan Chandra, Aniket Bera, and Dinesh Manocha.
\newblock Emotions don't lie: An audio-visual deepfake detection method using affective cues.
\newblock In \emph{ACM MM}, pages 2823--2832, 2020.

\bibitem[Narayan et~al.(2023)Narayan, Agarwal, Thakral, Mittal, Vatsa, and Singh]{narayan2023df}
Kartik Narayan, Harsh Agarwal, Kartik Thakral, Surbhi Mittal, Mayank Vatsa, and Richa Singh.
\newblock Df-platter: Multi-face heterogeneous deepfake dataset.
\newblock In \emph{CVPR}, pages 9739--9748, 2023.

\bibitem[Nightingale and Farid(2022)]{Nightingale2022}
Sophia~J. Nightingale and Hany Farid.
\newblock Ai-synthesized faces are indistinguishable from real faces and more trustworthy.
\newblock \emph{Proceedings of the National Academy of Sciences}, 119\penalty0 (8):\penalty0 e2120481119, 2022.

\bibitem[Nirkin et~al.(2019)Nirkin, Keller, and Hassner]{nirkin2019fsgan}
Yuval Nirkin, Yosi Keller, and Tal Hassner.
\newblock Fsgan: Subject agnostic face swapping and reenactment.
\newblock In \emph{ICCV}, pages 7184--7193, 2019.

\bibitem[{PBS NEWS}(2023)]{hongkong}
{PBS NEWS}.
\newblock Deepfakes in video group.
\newblock \url{https://www.channelnewsasia.com/commentary/deepfake-scam-video-conference-zoom-hong-kong-employee-4103266}, 2023.
\newblock Accessed: 2024-08-05.

\bibitem[Pellicer et~al.(2024)Pellicer, Li, and Angelov]{pellicer2024pudd}
Alvaro~Lopez Pellicer, Yi Li, and Plamen Angelov.
\newblock Pudd: Towards robust multi-modal prototype-based deepfake detection.
\newblock In \emph{CVPR}, pages 3809--3817, 2024.

\bibitem[Perov et~al.(2020)Perov, Gao, Chervoniy, Liu, Marangonda, Um{\'e}, Dpfks, Facenheim, RP, Jiang, et~al.]{perov2020deepfacelab}
Ivan Perov, Daiheng Gao, Nikolay Chervoniy, Kunlin Liu, Sugasa Marangonda, Chris Um{\'e}, Mr Dpfks, Carl~Shift Facenheim, Luis RP, Jian Jiang, et~al.
\newblock Deepfacelab: Integrated, flexible and extensible face-swapping framework.
\newblock \emph{arXiv preprint arXiv:2005.05535}, 2020.

\bibitem[Pidhorskyi et~al.(2020)Pidhorskyi, Adjeroh, and Doretto]{pidhorskyi2020adversarial}
Stanislav Pidhorskyi, Donald~A Adjeroh, and Gianfranco Doretto.
\newblock Adversarial latent autoencoders.
\newblock In \emph{CVPR}, pages 14104--14113, 2020.

\bibitem[Qian et~al.(2020)Qian, Yin, Sheng, Chen, and Shao]{thinking}
Yuyang Qian, Guojun Yin, Lu Sheng, Zixuan Chen, and Jing Shao.
\newblock Thinking in frequency: Face forgery detection by mining frequency-aware clues.
\newblock In \emph{ECCV}, pages 86--103, 2020.

\bibitem[Rodr{\'\i}guez et~al.(2017)Rodr{\'\i}guez, Cucurull, Gonfaus, Roca, and Gonzalez]{rodriguez2017age}
Pau Rodr{\'\i}guez, Guillem Cucurull, Josep~M Gonfaus, F~Xavier Roca, and Jordi Gonzalez.
\newblock Age and gender recognition in the wild with deep attention.
\newblock \emph{Pattern Recognition}, 72:\penalty0 563--571, 2017.

\bibitem[Rossler et~al.(2019)Rossler, Cozzolino, Verdoliva, Riess, Thies, and Nie{\ss}ner]{ffdata}
Andreas Rossler, Davide Cozzolino, Luisa Verdoliva, Christian Riess, Justus Thies, and Matthias Nie{\ss}ner.
\newblock Faceforensics++: Learning to detect manipulated facial images.
\newblock In \emph{ICCV}, pages 1--11, 2019.

\bibitem[Rothe et~al.(2015)Rothe, Timofte, and Van~Gool]{rothe2015dex}
Rasmus Rothe, Radu Timofte, and Luc Van~Gool.
\newblock Dex: Deep expectation of apparent age from a single image.
\newblock In \emph{ICCVW}, pages 10--15, 2015.

\bibitem[Sharma and Chakraborty(2024)]{sharma2024review}
Pavan~Kumar Sharma and Pranamesh Chakraborty.
\newblock A review of driver gaze estimation and application in gaze behavior understanding.
\newblock \emph{Engineering Applications of Artificial Intelligence}, 133:\penalty0 108117, 2024.

\bibitem[Sharma et~al.(2023)Sharma, Mir, and Singh]{sharma2023scale}
Vipal~Kumar Sharma, Roohie~Naaz Mir, and Chandrapal Singh.
\newblock Scale-aware cnn for crowd density estimation and crowd behavior analysis.
\newblock \emph{Computers and Electrical Engineering}, 106:\penalty0 108569, 2023.

\bibitem[Shehnepoor et~al.(2022)Shehnepoor, Togneri, Liu, and Bennamoun]{shehnepoor2022spatio}
Saeedreza Shehnepoor, Roberto Togneri, Wei Liu, and Mohammed Bennamoun.
\newblock Spatio-temporal graph representation learning for fraudster group detection.
\newblock \emph{TNNLS}, 2022.

\bibitem[Shen et~al.(2020)Shen, Gu, Tang, and Zhou]{shen2020interpreting}
Yujun Shen, Jinjin Gu, Xiaoou Tang, and Bolei Zhou.
\newblock Interpreting the latent space of gans for semantic face editing.
\newblock In \emph{CVPR}, pages 9243--9252, 2020.

\bibitem[Shiohara and Yamasaki(2022)]{shiohara2022detecting}
Kaede Shiohara and Toshihiko Yamasaki.
\newblock Detecting deepfakes with self-blended images.
\newblock In \emph{CVPR}, pages 18720--18729, 2022.

\bibitem[Smith et~al.(2013)Smith, Yin, Feiner, and Nayar]{smith2013gaze}
Brian~A Smith, Qi Yin, Steven~K Feiner, and Shree~K Nayar.
\newblock Gaze locking: passive eye contact detection for human-object interaction.
\newblock In \emph{ACM Symposium on UIST}, pages 271--280, 2013.

\bibitem[Tan et~al.(2023)Tan, Wang, Wang, Yang, Chen, and Guo]{tan2023deepfake}
Lingfeng Tan, Yunhong Wang, Junfu Wang, Liang Yang, Xunxun Chen, and Yuanfang Guo.
\newblock Deepfake video detection via facial action dependencies estimation.
\newblock In \emph{AAAI}, pages 5276--5284, 2023.

\bibitem[Tian et~al.(2024)Tian, Chen, Zhou, and Hu]{tian2024illumination}
Kaiyue Tian, Chen Chen, Yichao Zhou, and Xiyuan Hu.
\newblock Illumination enlightened spatial-temporal inconsistency for deepfake video detection.
\newblock In \emph{ICME}, pages 1--6. IEEE, 2024.

\bibitem[Tiktok(2025)]{trumpZ}
Tiktok.
\newblock Fake ai videos about trump fights with zelenskyy.
\newblock \url{https://vt.tiktok.com/ZSMC4jW6g/}, 2025.
\newblock Accessed: 2025-03-07.

\bibitem[Vaswani et~al.(2017)Vaswani, Shazeer, Parmar, Uszkoreit, Jones, Gomez, Kaiser, and Polosukhin]{vaswani2017attention}
Ashish Vaswani, Noam Shazeer, Niki Parmar, Jakob Uszkoreit, Llion Jones, Aidan~N Gomez, {\L}ukasz Kaiser, and Illia Polosukhin.
\newblock Attention is all you need.
\newblock \emph{NeurIPS}, 30, 2017.

\bibitem[Wang et~al.(2021)Wang, Gao, Lin, and Yuan]{wang2021pixel}
Qi Wang, Junyu Gao, Wei Lin, and Yuan Yuan.
\newblock Pixel-wise crowd understanding via synthetic data.
\newblock \emph{IJCV}, 129\penalty0 (1):\penalty0 225--245, 2021.

\bibitem[Wang and Chow(2023)]{wang2023noise}
Tianyi Wang and Kam~Pui Chow.
\newblock Noise based deepfake detection via multi-head relative-interaction.
\newblock In \emph{AAAI}, pages 14548--14556, 2023.

\bibitem[Wang et~al.(2017)Wang, Ali, and Angelov]{wang2017gender}
Xiaofeng Wang, Azliza~Mohd Ali, and Plamen Angelov.
\newblock Gender and age classification of human faces for automatic detection of anomalous human behaviour.
\newblock In \emph{CYBCONF}, pages 1--6, 2017.

\bibitem[Waseem et~al.(2023)Waseem, Bakar, Ahmed, Omar, and Eisa]{waseem2023deepfake}
Saima Waseem, Syed Abdul Rahman Syed~Abu Bakar, Bilal~Ashfaq Ahmed, Zaid Omar, and Taiseer Abdalla~Elfadil Eisa.
\newblock Deepfake on face and expression swap: A review.
\newblock \emph{IEEE Access}, 11:\penalty0 117865--117906, 2023.

\bibitem[Xie et~al.(2024)Xie, Jiao, Wu, Guo, and Hong]{xie2024active}
Zhao Xie, Chang Jiao, Kewei Wu, Dan Guo, and Richange Hong.
\newblock Active factor graph network for group activity recognition.
\newblock \emph{TIP}, 2024.

\bibitem[Xu et~al.(2023)Xu, Liang, Jia, Yang, Zhang, and He]{xu2023tall}
Yuting Xu, Jian Liang, Gengyun Jia, Ziming Yang, Yanhao Zhang, and Ran He.
\newblock Tall: Thumbnail layout for deepfake video detection.
\newblock In \emph{ICCV}, pages 22658--22668, 2023.

\bibitem[Yuan et~al.(2021)Yuan, Ni, and Wang]{yuan2021spatio}
Hangjie Yuan, Dong Ni, and Mang Wang.
\newblock Spatio-temporal dynamic inference network for group activity recognition.
\newblock In \emph{ICCV}, pages 7476--7485, 2021.

\bibitem[Zhang et~al.(2024)Zhang, Qi, Wang, Li, and Lyu]{zhang2024comics}
Cong Zhang, Honggang Qi, Shuhui Wang, Yuezun Li, and Siwei Lyu.
\newblock Comics: End-to-end bi-grained contrastive learning for multi-face forgery detection.
\newblock \emph{TCSVT}, 2024.

\bibitem[Zhang et~al.(2022)Zhang, Liu, and Lu]{zhang2022gazeonce}
Mingfang Zhang, Yunfei Liu, and Feng Lu.
\newblock Gazeonce: Real-time multi-person gaze estimation.
\newblock In \emph{CVPR}, pages 4197--4206, 2022.

\bibitem[Zhou et~al.(2017)Zhou, Han, Morariu, and Davis]{zhou2017two}
Peng Zhou, Xintong Han, Vlad~I Morariu, and Larry~S Davis.
\newblock Two-stream neural networks for tampered face detection.
\newblock In \emph{CVPRW}, pages 1831--1839. IEEE, 2017.

\bibitem[Zhou et~al.(2021)Zhou, Wang, Liang, and Shen]{zhou2021face}
Tianfei Zhou, Wenguan Wang, Zhiyuan Liang, and Jianbing Shen.
\newblock Face forensics in the wild.
\newblock In \emph{CVPR}, pages 5778--5788, 2021.

\bibitem[Zhuang et~al.(2020)Zhuang, Ni, Xu, Yang, Zhang, Li, and Gao]{MUGGLE}
Ning Zhuang, Bingbing Ni, Yi Xu, Xiaokang Yang, Wenjun Zhang, Zefan Li, and Wen Gao.
\newblock Muggle: Multi-stream group gaze learning and estimation.
\newblock \emph{TCSVT}, 30\penalty0 (10):\penalty0 3637--3650, 2020.

\end{thebibliography}
}

\end{document}